\documentclass[conference]{IEEEtran}                                                          

\IEEEoverridecommandlockouts                              

\usepackage{graphicx}
\usepackage{caption}
\usepackage{subcaption}
\usepackage{color}
\usepackage{amsmath,amssymb}
\usepackage[]{algorithm2e}

\newcommand{\appropto}{\mathrel{\vcenter{
  \offinterlineskip\halign{\hfil$##$\cr
    \propto\cr\noalign{\kern2pt}\sim\cr\noalign{\kern-2pt}}}}}
\newcommand{\realspace}{\mathbb{R}}

\newcommand{\gaussdist}[3]{{\cal N}_{#1}(#2,#3)} 
\newcommand{\gaussmix}[6]{ \sum_{#5=1}^{#6}{ #4_{#5} \gaussdist{#1}{#2_{#5}}{#3_{#5}} } }
\newcommand{\esoftmaxfuncbias}[6]{\frac{e^{ #2_{#1}^T #3 + #6_{#1}}}{\sum_{#4=1}^{#5}{e^{ #2_{#4}^T #3 + #6_{#4}}}}} 
\newcommand{\RealSpace}[1]{\mathbb{R}^{#1}}

\newcommand{\pareqref}[1]{(\ref{eq:#1})}

\title{Closed-loop Bayesian Semantic Data Fusion \\ for Collaborative Human-Autonomy Target Search}

\author{Luke Burks, Ian Loefgren, Luke Barbier, Jeremy Muesing, Jamison McGinley, Sousheel Vunnam, and Nisar Ahmed$^*$
\thanks{$^*$Ann And H.J. Smead Aerospace Engineering Sciences Department, 429 UCB, University of Colorado Boulder, Boulder CO 80309, USA. E-mail:{\ttfamily[luke.burks;nisar.ahmed@colorado.edu}.}
}

\begin{document}

\maketitle

\begin{abstract}
In search applications, autonomous unmanned vehicles must be able to efficiently reacquire and localize mobile targets that can remain out of view for long periods of time in large spaces. As such, all available information sources must be actively leveraged -- including imprecise but readily available semantic observations provided by humans. 
To achieve this, this work develops and validates a novel collaborative human-machine sensing solution for dynamic target search. Our approach uses continuous partially observable Markov decision process (CPOMDP) planning to generate vehicle trajectories that optimally exploit imperfect detection data from onboard sensors, as well as semantic natural language observations that can be specifically requested from human sensors. The key innovation is a scalable hierarchical Gaussian mixture model formulation for efficiently solving CPOMDPs with semantic observations in continuous dynamic state spaces. The approach is demonstrated and validated with a real human-robot team engaged in dynamic indoor target search and capture scenarios on a custom testbed.    
\end{abstract}

\section{Introduction}
Dynamic target search and localization remains a very active research area for unmanned autonomous vehicle systems. 
Solutions typically leverage joint state space models of target dynamics, mobile sensor platform motion, and sensor observations to solve challenging combined optimal control and estimation problems. 
However, practical algorithms for data fusion and decision making can still be too computationally expensive and brittle to ensure full vehicle autonomy. 

In many cases, human operators and users can act as `human sensors' that contribute valuable information beyond the reach of autonomous vehicle sensors. For instance, operators in search and tracking missions using small unmanned aerial systems (UAS) can provide `soft data' to narrow down possible survivor locations using semantic natural language observations (e.g. `Nothing is around the lake'; `Something is moving towards the fence'), or provide estimates of physical quantities (e.g. masses/sizes of obstacles, distances from landmarks) to help autonomous vehicles better understand search areas and improve online decision making with limited computational resources. This naturally raises the question of how autonomous reasoning can actively and opportunistically engage human reasoning to improve their own performance. 

\begin{figure}[t]
\centering	\includegraphics[width=0.51\textwidth]{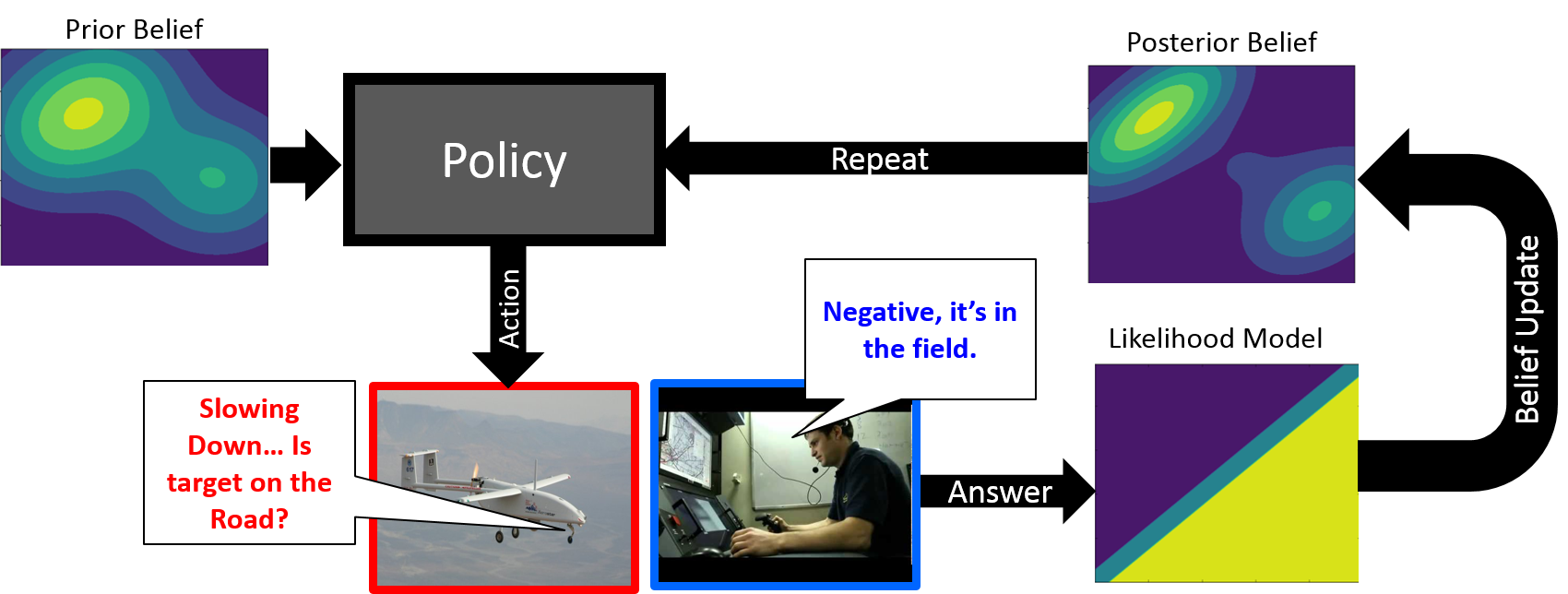}
	\caption{{\scriptsize Closed-loop collaborative Bayesian target search using a non-myopic policy for simultaneous semantic querying and sensor vehicle motion planning.}}
    \label{fig:infExamples}
    \vspace{-0.2 in}
\end{figure}

We present a rigorous framework for intelligent human-autonomy interaction that not only leverages combined robot-human sensing, but is also tightly integrated with dynamic platform decision making and planning. 
Our approach uses Bayesian data fusion to exploit soft data with minimal effort on the part of human sensors and autonomous robotic sensor platforms. Such `plug and play' human sensing for robot state estimation was explored in \cite{Kaupp2007, Bourgault2008} for restricted types of human observations, and has received increased attention in recent years \cite{Khaleghi10,Dani2014}. 
In this paper, we combine our recent work on Bayesian semantic natural language human data fusion \cite{Ahmed-TRO-2013,Sweet2016} with concepts from optimal active sensing, in order to develop new methods for \textit{interactive} human-robot semantic sensing. Here we focus on the challenging problem of non-myopic decision making for \textit{simultaneous (tightly coupled) vehicle motion planning and human sensor querying} in continuous dynamic search environments. 
As shown in Figure \ref{fig:infExamples}, our approach leads to joint action-query \textit{policies} (i.e. control laws). The policies tell the robot how to respond to target location uncertainty, so that it simultaneously makes optimal decisions about how to move/sense on its own in the environment and about which semantic natural language questions it should ask human sensors in order to `pull' useful information. The human only needs to act as a (voluntary) sensor, and does not need to actively control or plan for the robot; furthermore, the policy lets the robot conduct an optimal search with complex non-Gaussian uncertainties, even without human input. 

Our technical approach builds on recent foundational work for efficiently finding policies based on continuous partially observable Markov decision process (CPOMDP) models \cite{Burks2017}. While this CPOMDP approach provides several nice theoretical features for collaborative dynamic target search, we address some open issues that are crucial for real practical system deployment. In particular, we present a scalable hierarchical CPOMDP solution that allows our framework to be deployed in arbitrarily complex environments, e.g. large indoor settings with multiple rooms/search areas, many possible semantic grounding references, and moving targets, which would otherwise be intractable for a single CPOMDP policy to handle. We also present test demonstration results that validate our approach with a real human-robot team engaged in dynamic indoor target search and capture scenarios. While our presentation is grounded in dynamic target search problems, our human-autonomy collaboration framework can be applied to other interactive dynamic data fusion problems as well.

This paper is structured as follows. Section II reviews background and related work. Section III presents our new hierarchical CPOMDP framework for optimal search and interactive semantic soft data querying, in the context of indoor dynamic target search. Section IV provides demonstration results on our custom human-robot team target search testbed, and Section V presents conclusions and future work. 



\section{Background and Related Work}

In this work, we focus on search and localization problems where the number and type of targets are known, and where mobile sensor platform dynamics and observation models are known. Several fundamental difficulties become apparent even in this setting, which lead to brittleness in practice. 
Firstly, vehicles are subject to constraints on motion, size, weight, power, and cost; this limits their computing and sensing payloads as well as their operating time and range. 
Secondly, the sensing and planning horizons for approximate optimal search algorithms are inherently limited. This not only restricts the ability to correctly detect and sense targets, but also the ability to execute adaptive long term information gathering strategies in complex dynamic environments.  
Finally, sensing platforms may only have access to imperfect/highly uncertain target behavior models. This can lead to non-Gaussian probability distributions over target states, and make online planning and sensing/data fusion even more difficult. 
 
Formal integration of robotic and human perception can greatly improve the efficiency and robustness of autonomous decision making, especially in situations where uncertainties cannot be well-characterized in advance and must be adapted on the fly. 
Soft data can be broadly related to either `abstract' phenomena that cannot be measured by robotic sensors (e.g. labels for occupied/unoccupied rooms, object categories and behaviors) or measurable dynamical physical states that must be monitored continuously (object position, velocity, attitude, temperature, size, mass, etc.) \cite{HallBook}. We examine the problem of \textit{active soft data fusion}, and build on methods for addressing the following key issues: (i) soft semantic data modeling; and (ii) active semantic sensing for intelligent planning. 

\subsection{Mixture-based Bayesian Soft Data Fusion}
Ref. \cite{Ahmed-TRO-2013} showed how to model and fuse flexible semantic natural language data to provide a broad range of positive/negative information for Bayesian state estimation, e.g. `The target is parked near the tree in front of you', `Nothing is next to the truck heading North'. This fusion algorithm directly plugs into Gaussian mixture filters for robotic state estimation, which can accurately represent complex posterior pdfs while avoiding the curse of dimensionality. 
Suppose $s_k \in \mathbb{R}^n$ is a random vector representing some continuous state of interest at discrete time $k$ (e.g.\ target location, velocity, heading) with prior pdf $p(s_k)$, which may already be conditioned on hard/soft sensor data and predicted forward in time from according to known a stochastic state transition pdf via the Chapman-Kolmogorov equation. Let let $D_k$ be a discrete random variable representing a human-generated semantic observation related to $s_k$. 
Bayes' rule gives the posterior pdf
{
\allowdisplaybreaks
\abovedisplayskip = 2pt
\abovedisplayshortskip = 1pt
\belowdisplayskip = 0pt
\belowdisplayshortskip = 0pt
\begin{align}\label{eq:bayes_update}
p(s_k | D_k = i) = \frac{P(D_k=i| s_k)p(s_k)}{\int P(D_k=i| s_k)p(s_k)ds_k}
\end{align}
}
where the likelihood function $P(D_k|s_k)$ captures the human's semantic classification behavior conditioned on the true state $s_k$. If $D_k= i$ corresponds to one of $m$ exclusive semantic categories for a known dictionary of state observations, then a \emph{softmax function} (i.e. multinomial logistic function) can be used to model $P(D_k=i|s_k)$,
{
\allowdisplaybreaks
\abovedisplayskip = 2pt
\abovedisplayshortskip = 1pt
\belowdisplayskip = 0pt
\belowdisplayshortskip = 0pt
\begin{align}\label{eq:softmax}
P(D_k = i| s_k) = \esoftmaxfuncbias{i}{w}{s_k}{j}{m}{b} 
\end{align}
}
where $w_j$ and $b_j$ are vector weight and scalar bias for class label $i$. For a sufficiently rich dictionary of semantic observations $D_k$, multiple softmax models can be defined via eq. \pareqref{softmax} with $m=2$ for different binary sets of semantically similar class labels (`nearby' vs. `not nearby', `next to' vs. `not next to', `close by' vs. `not close by', etc.), so that they need not be treated as mutually exclusive labels within a single large softmax model. 
The likelihood parameters $w_j$ and $b_j$ can be learned from semantic human sensor calibration data \cite{Ahmed-TRO-2013} and algebraically manipulated to shift, dilate, rotate, and geometrically constrain semantic class boundaries in $\mathbb{R}^{n}$ \cite{Sweet2016}. 

Eq. \pareqref{bayes_update} must be approximated for recursive Bayesian data fusion with softmax likelihoods, since the exact posterior pdf $p(s_k|D_k)$ cannot be obtained in closed-form for any $p(s_k)$. If $P(D_k = i|s_k)$ is generally given by a softmax model for observation label $i$ and the prior is given by a finite Gaussian mixture (GM) with $m_p$ prior components,
{
\allowdisplaybreaks
\abovedisplayskip = 2pt
\abovedisplayshortskip = 2pt
\belowdisplayskip = 1pt
\belowdisplayshortskip = 1pt
\begin{align}
p(s_k) = \gaussmix{x_k}{\mu}{\Sigma}{w}{p}{m_p} \nonumber
\end{align}
}
(where $w_p$, $\mu_p \in \RealSpace{n} $, and $\Sigma_p \in \RealSpace{n \times n} $ are the weights, mean vector and covariance for mixand $p$), then $p(s_k|D_k=i)$ can be well-approximated by an $m_p$ component GM,
{
\allowdisplaybreaks
\abovedisplayskip = 2pt
\abovedisplayshortskip = 2pt
\belowdisplayskip = 1pt
\belowdisplayshortskip = 1pt
\begin{align}
p(s_k|D_k = i) \approx \gaussmix{x_k}{\mu}{\Sigma}{w}{q}{m_p}. \label{eq:post_mixture}
\end{align}
}
The weights, means and covariances of posterior component $q$ can be determined by fast numerical approximations methods \cite{Ahmed-TRO-2013}, and mixture compression methods can be used to manage the growth of mixture terms due to non-linear dynamics or application of non-convex `multimodal' softmax models \cite{Runnalls-AES-2007}. 

\subsection{Active Semantic Sensing for Planning Under Uncertainty}
A major challenge for problems like target localization is that dynamics and uncertainties can quickly become quite non-linear and non-Gaussian, given the types of semantic information available for fusion (e.g. negative information from `no detection' readings \cite{Koch04}). 
As a result, typical stovepiped approaches to control/planning and sensing/estimation can lead to poor performance, since they rely on overly simplistic uncertainty assumptions. 
Constraints on human and robot performance also place premiums on when and how often collaborative data fusion can occur. 
%
This can be cast as an active sensing problem where \emph{value of information} (VOI) must be evaluated \cite{Kaupp2010}. Put loosely, this implies that costly but valuable sensor observations should only be collected if (no matter their outcome) they lead to an improvement in utility for decision making under uncertainty. 
%




Target search problems in uncertain environments can be cast as Partially Observable Markov Decision Processes (POMDPs) to non-myopically integrate VOI-based reasoning. In general, POMDPs solvers seek a \textit{policy}, which maps a belief over the state space (i.e. a pdf) to a recommended action. These actions seek to maximize the expected time-discounted reward over time. Exact solutions to POMDPs are impractical in all but the most trival of problems, and a variety of approximate solutions have been proposed. One class of POMDP approximation known as Point-Based Value Iteration \cite{Pineau2003} and various related algorithms \cite{Spaan2005} \cite{Kurniawati2008} rely on recursively solving the corresponding Bellman equations with known observation and transition models for some subset of possible beliefs that might be encountered during policy execution. When specified over discrete state, observation, and action spaces, the computational complexity explodes quickly with the number of joint configurations across each space. 

The introduction of the CPOMDP method \cite{Porta2006} showed that sets of Gaussian Mixture (GM) models can be used to approximate the policy over a continuous state space, while Switching-Mode POMDPs \cite{Brunskill2010} further extended the CPOMDP framework to account for non-constant transition functions such as those caused by the presence of obstructions in the space. Finally, Variational Bayes POMDPs (VB-POMDPs) \cite{Burks2017} were developed to handle non-Gaussian observation models in the form of softmax models, which easily model semantic observation statements and excel at parsing  semantic input statements over a continuous space while significantly decreasing the computational cost of finding and implementing a policy. 


In many realistic applications, it is impractical to construct the problem as a single POMDP. While continuous state spaces address the issue of large state spaces, they do not naturally account for discontinuous transitions, e.g. such as those involving obstacles for mobile platform motion planning. While this was addressed by the Switching-Mode POMDP framework \cite{Brunskill2010}, the presence of many objects and/or separate rooms can add prohibitively more computation to the policy solution. Having large numbers of objects which can act as anchors for semantic human sensor observations also leads to a large observation space, which also  increases computational costs. Even with recent advances in scalability such as those showcased by VB-POMDP, existing policy approximation methods are ill-suited to handle the kind of complex and informationally dense settings found in real world applications.  

Consider the problem of tracking a target through a typical indoor environment, which we refer to here as the `Cops and Robbers' (CNR) problem. The primary robotic agent, referred to here as the cop, is tracking the robber with the aid of a human viewing the scene through a series of security cameras. With its separate but connected rooms, and information-dense environment filled with objects that can be referenced for observations, this scenario presents a challenge for typical POMDP approaches. CPOMDP methods struggle with the number and complexity of switching-modes required to encapsulate objects and walls. In principle, online POMDP solvers could be used to come up with acceptable approximations for such problems using policy search techniques. But online solvers struggle with problems like CNR, since rewards can only be obtained at a single point in the state space, i.e. when the robber is caught.  This causes problems for large state spaces as positive reward states will often lie beyond an online solver's effective planning horizon, and thus intermediate rewards cannot be obtained to promote adequate policy exploration. However, since the continuous indoor environments is easily separable into connected regions, this suggests exploring the use of multiple connected CPOMDP policies that can be obtained offline.

\section{Hierarchical Continuous POMDPs}


We address this scalability problem in the context of the CNR problem for target search in complex indoor search spaces, e.g. see Fig. \ref{fig:maps}. Our approach is to find a separate continuous POMDP policy for each distinct room in a particular map, where obstacles are sparse enough not to necessitate the switching modes used in \cite{Brunskill2010}. Each of these room level policies is then treated as an action selection by a discrete POMDP policy over the rooms. This leads to a novel hierarchical CPOMDP policy that can not only take fuse low-level semantic soft information about target locations in metric physical space (e.g. `next to the chair'; `not by the cooler'), but also exploit higher-level semantic data about target locations in abstract label spaces, i.e. room designations (`in the kitchen'; `not in the dining room'). By accounting for the dependencies between these different types of high-level and low-level semantic data, we arrive at an intelligent hierarchical decision making policy that enables top-down motion planning (i.e. determine which areas to search, and then how to search them), as well as determination of the best set of high-level and low-level semantic queries for a human sensor that will ensure rapid capture of the robber. 


\subsection{Lower Level CPOMDP}
The lower level CPOMDP for each room is specified over a state $ S = \realspace^{4}$, which consists of two bounded continuous random variables for each agent at each time step $t$. The cop's state variables are changed deterministically with actions while the robber's are assumed to be a Gaussian random walk. The cop can choose from among 5 noisy movement actions $A_{m} =\{East,West,North,South,Stay\}$, and can ask questions about the robber's spatial relation to each object in the room such that $A_{q} = \{{Objects}\} \times \{Left,Right,Front,Behind\}$. The full discrete action space is then $A = A_{m} \times A_{q}$. An example action might be "Move East and ask 'Is the robber in front of the fern?'".

The cop relies on two sources of observations. First is the viewcone, which supplies binary observations in the form of $O_{v} = \{Detection, No Detection\}$, depending on whether the robber is physically located within the specified viewcone. In this implementation, the viewcone is approximated by a 1 meter square box centered on the cops position. Including orientation into the state vector would remove the need for this approximation, but further increase the dimensionality of the state. Secondly, the cop receives answers to it's questions $A_{q}$, in the binary form $O_{q} = \{Yes,No\}$. The full discrete observation space is given as $O = O_{v} \times O_{q}$, with a set size $|O| = 4$. In this work continuous space observation models were constructed using softmax models as in \cite{Burks2017}. 

The reward function is specified by an action dependent Gaussian mixture function (i.e. which can be thought of as an unnormalized GM pdf that may contain negative mixing weights). Rewards are given for the co-location of the cop and the robber to within a meter of each other, with the mean of each mixand shifted by the cops per action movement distance. This is to further encourage the cop to take an action directly toward the robber, which effectively improves the viewcone approximation mentioned above by increasing the chance the robber's position in the approximated viewcone will fall within the real viewcone.

\begin{figure}
      \centering	
      \begin{subfigure}[h]{.15\textwidth}
          \includegraphics[width=\textwidth]{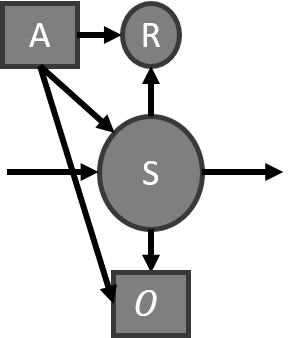}
          \caption{\scriptsize Standard POMDP}
          \label{fig:standModel}
      \end{subfigure}
      ~
      \begin{subfigure}[h]{.15\textwidth}
          \includegraphics[width=\textwidth]{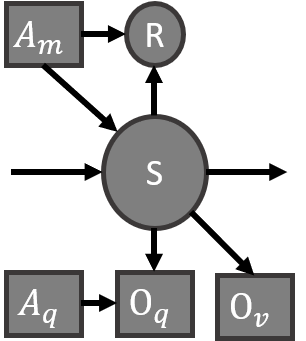}
          \caption{\scriptsize Factored POMDP}
          \label{fig:factModel}
      \end{subfigure}
      \caption{\scriptsize POMDP graphical models with different action/observation factorizations.}
      \label{fig:lowerModels}
      \vspace{-0.2in}
\end{figure}

In describing the action and observations spaces of the problem for implementation it would be perfectly valid to solve the CPOMDP over the combined action and observation space without mentioning that each is factored into two distinct sets. However, this increases the difficulty of  implementation when trying to account for the diverse results of the combined action/observation. In this case, the cop's movement actions $A_{m}$ primarily effect the state without changing the observations, while the cop's `question actions' $A_{q}$ have no effect on the state at the current time, and fully dictate the meaning of the observations $O_{q}$. Similarly, the viewcone observations $O_{v}$ are only state dependent and thus independent of either action, while $O_{q}$ depends on both state and action. Factoring each space into it's constituent parts allows for simpler handling of these dependencies, and increases the explainability of the cop's actions and the changes in it's beliefs. The differences in the two approaches are summed up in Figure \ref{fig:lowerModels}. 

\subsection{Higher Level Discrete POMDP}

The higher level discrete POMDP is specified on a state vector consisting of all rooms. The state represents the robber's current position, and the robber randomly transitions according to the particular connections between rooms in the map being used. The cops position is not represented in the state vector. The cop can choose movement actions $A_{m}$ corresponding to each room, which will deterministically move the cop to that room, as well as questions actions $A_{q}$, which will ask the human if the robber is in a particular room. As with lower level policies, the full action space is $A = A_{m} \times A_{q}$. An example action would be ``Search the Library and ask 'Is the robber in the Kitchen?'". 

In the higher level discrete POMDP, the cop is rewarded for choosing to move to the room the containing the robber, and penalized for choosing the wrong room. The cop receives viewcone observations $O_{v}$ at each time step, similar to the lower level CPOMDP. Given that being in the same room as the robber does not guarantee a viewcone detection, likelihoods for $O_{v} = Detection$ are fairly low for any given time step. The cop can also receive responses to it's questions $A_{q}$ in the form of human observations $O_{q}$.

We use the Point-Based Value Iteration approach from \cite{Pineau2003} to find the policy for the discrete layer. 

\subsection{Hierarchy and Question Lists}

At each time step, the policy chooses an action consisting of a room to search and a room to query. If the cop is outside the search room, it is directed to go there. Otherwise, if the cop is already in the search room, the lower level CPOMDP policy is queried to provide a movement action. The query room is asked about in the form``Is the robber in (room)?", and low level CPOMDP policy for that room gives an additional question about the robber's relation to an object in that room. 

In this implementation the human sensor receives a question from the cop at every time step. Because the policy was trained to expect responses from the human sensor, steps were the human fails to answer are unknown events from the system's standpoint, i.e. they are not accounted for when solving for the policy. One method for handling these failures would be to include a``Null" observation with a uniform likelihood across states to represent a lack of human observation. Further steps could be taken by incorporating a form of human attention model into the state vector and an option to ask a ``Null" question when the policy believes the human would not be able to answer.

In most applications, the desired result of a POMDP query is the action with the highest value for the current belief. This makes sense in most contexts as only one action can be taken at a time. However, in our problem multiple questions could be displayed to the human at each time step, and so we want to ask the $N$ most valuable questions. In both discrete and continuous policies using PBVI-type approximations each policy element, or $\alpha$-element, contained in $\Gamma$ corresponds to an action and encodes part of the approximate value function over beliefs. As each $\alpha$-element is specified over the entire belief space, it can provide a value for its action at any belief, even were it does not provide the maximum value. Therefore, the $\alpha$-elements with the top $N$ values can be said to correspond to the top $N$ actions. As multi $\alpha$-elements might correspond to the same action, this does not guarantee $N$ unique actions. However, as all $\alpha$-elements must be evaluated to choose the correct action for a belief, the top $N$ unique actions can still be chosen. This also implies that choosing a list of actions requires only the minimal extra computation of a sorting function substituted for an argmax, as in Algorithm 1. 

\newcommand\tab[1][1cm]{\hspace*{#1}}

\begin{algorithm}
    \SetKwInOut{Input}{Input}
    \SetKwInOut{Output}{Output}
    \underline{Choose $N$ actions}\\
    \Input{$b(s)$, $\Gamma$, $N$}
    \textbf{for} $\forall \alpha \in \Gamma$:\\
    \tab V($\alpha$) = $\int \alpha(s) b(s) ds$ \\
    list = sort(V) \\
    return list[0:N]\\
    
    \caption{{\scriptsize Choose the best $N$ actions recommended by the policy.}}
\end{algorithm}

\subsection{Belief Updates}
The cop maintains a belief over the lower level continuous state using a GM pdf. This is converted to a discrete belief over rooms by carrying out a hard classification of each mixand to whichever room it's mean is contained in for the robber state dimensions. The belief for each room $r$ then becomes the sum of the weights of the mixands assigned to it.




\subsection{Dynamic Target Models}
The CPOMDP framework shown in \cite{Porta2006} is equipped to handle transition functions which can be modeled as Gaussian distributions with their mean shifted by the actions $\Delta a$ expected effect on the state $s$: 
\begin{align}
p(s'|s,a) = \phi(s'|s+\Delta a, \Sigma_{a})
\end{align}
When only a limited number of state components are directly controllable, the others are forced to execute a Gaussian random walk with the given variance. This prevents the use of target trajectories in search problems such as the one described here. In order to incorporate target dynamics, it is desirable to have a transition function of the form,
\begin{align}
p(s'|s,a) = \phi(s'|As+\Delta a, \Sigma_{a})
\end{align}
where $A$ is the state transition matrix which encapsulates changes in the state independent of actions. Bellman backups can be easily resolved with this alteration in the CPOMDP framework, thus permitting solutions to a broad class of continuous space planning problems.


\section{Application Demonstration}

Hierarchical CPOMDPs were implemented and tested on the Cops and Robots (CNR) Hardware platform at the Univeristy of Colorado at Boulder's Research and Engineering Center for Unmanned Vehicles. Individual agents playing the part of the cop and robber were instantiated on Turtlebots running from an Odroid U3 microcontroller on iRobot Create platforms. Cops and Robots is a physical simulation of a home environment, with semantic labels assigned to the rooms and objects within. These labels are known to both the cop and human, which allows for communication of information through a fixed codebook of possible observations.

\begin{figure}[t!]
\centering	\includegraphics[width=0.45\textwidth]{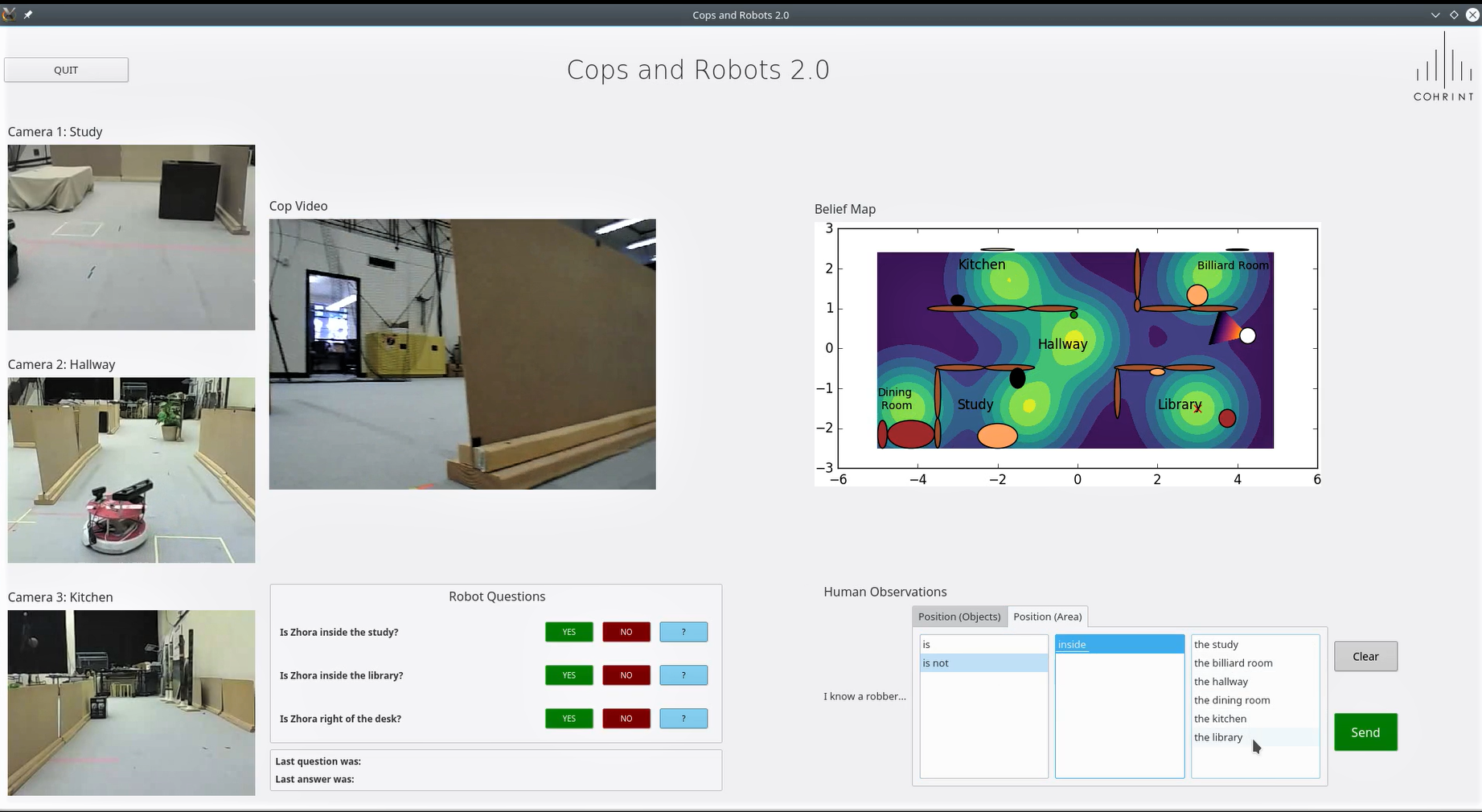}
	\caption{{\scriptsize Cops and Robots user interface: `robot pull' queries are answered in the lower middle panel with `Yes/No' buttons; voluntary `human push' sensor inputs are provided with the structured text input on the lower right panel.}}
    \label{fig:CNRInterface}
    \vspace{-0.3 in}
\end{figure}
The human interface, shown in Figure \ref{fig:CNRInterface}, visualizes the cop's belief about the robber's position as a heatmap, as well as the cop's position and viewcone detection range. The interface also displays a real-time feed from the cop's camera and various security cameras placed throughout the space. The security cameras each allow the human a fixed view of a room, while the cops camera facilitates observations in the cop's immediate area as well as a visual robber detection system. The human plays the role of a sensor, voluntarily passing information to the cop through the semantic codebook embedded in the interface and answering binary `yes/no' questions passed from the cop (e.g. `Is the robber in the kitchen?'; `Is robber in front of fern?'). The human is also required to validate visual detections of the robber, where a correct validation leads to successful capture of the robber. 

\begin{figure}[h]
	\centering	
	\begin{subfigure}[t]{.4\textwidth}
		\includegraphics[width=.85\textwidth]{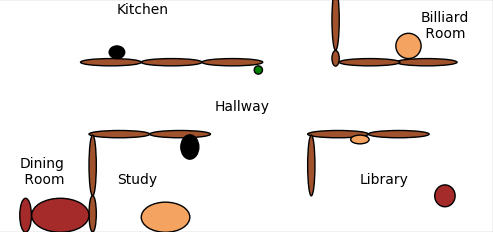}
		\caption{{\scriptsize First/familiar map}}
    	\label{fig:layoutMapA}
    \end{subfigure}
    
    \begin{subfigure}[t]{.4\textwidth}
    	\includegraphics[width=.85\textwidth]{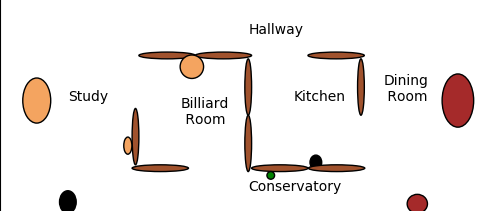}
		\caption{{\scriptsize Second/unfamiliar map}}
   		\label{fig:layoutMapC}
    \end{subfigure}
    \caption{\scriptsize{ Layouts for first (above) and second (below) maps}}
    \label{fig:maps}
\vspace{-0.05 in}
\end{figure}

The Hierarchical CPOMDP method was tested on two CNR maps, each with a different rooms structure. The first map, shown in \ref{fig:layoutMapA}, consisted primarily of a hallway running the length of the space, with rooms branching off on both sides. The second map, shown in \ref{fig:layoutMapC}, had the rooms in a semi-bipartite arrangement, with two sets connected through a long hallway and conservatory on the margins. In data collection, the human participant was fully familiarized with the first map beforehand, while the second map was presented as a previously unknown environment. 

\begin{figure}[t]
	\centering	
	\begin{subfigure}[t]{.2\textwidth}
		\includegraphics[width=1\textwidth]{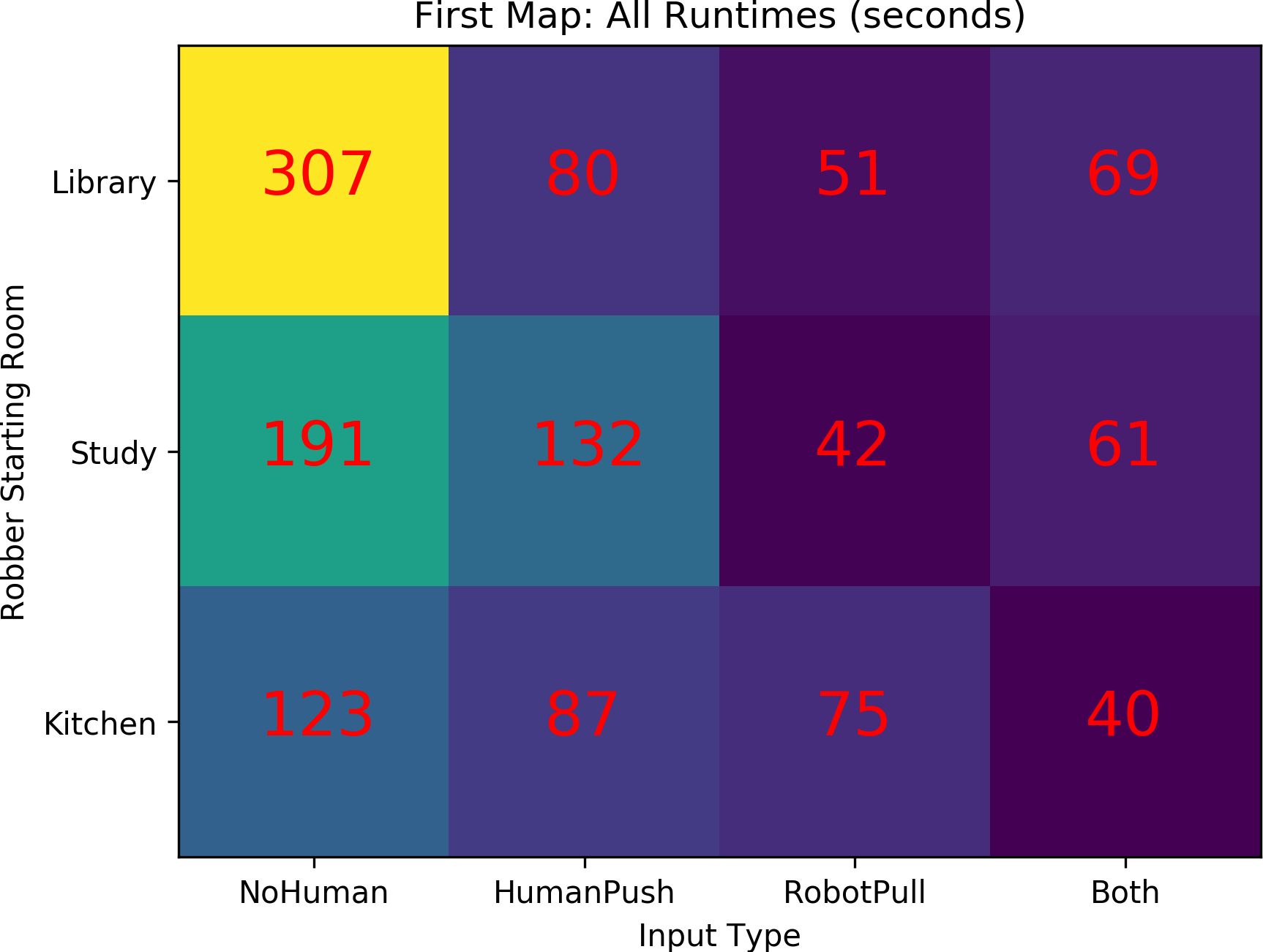}
		\caption{}
    	\label{fig:timesMapA}
    \end{subfigure}
    ~
    \begin{subfigure}[t]{.2\textwidth}
    	\includegraphics[width=1.1\textwidth]{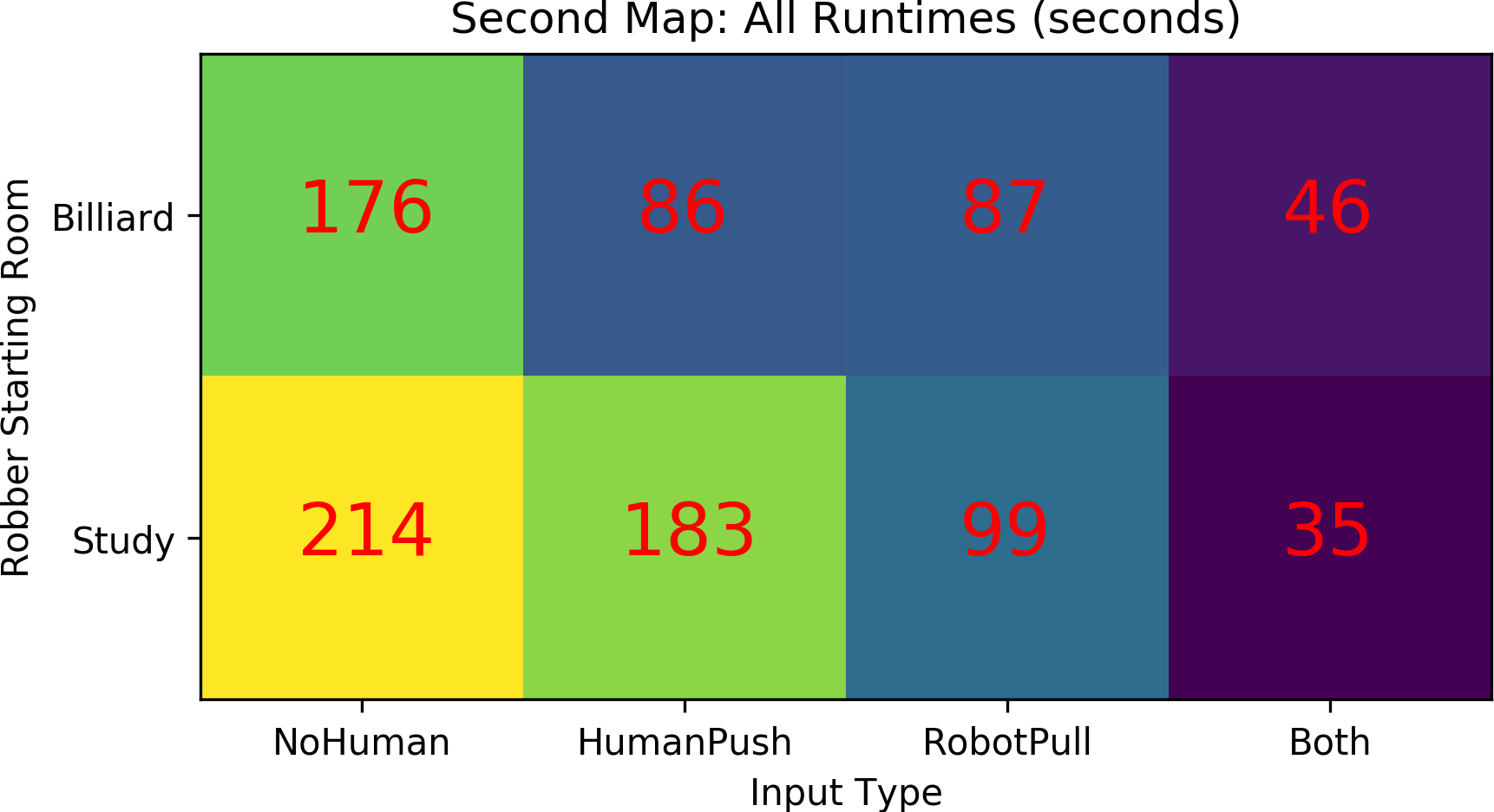}
		\caption{}
   		\label{fig:timesMapC}
    \end{subfigure}
    \caption{{\scriptsize Run times for first (a) and second (b) maps.}}
    \label{fig:allCatchTimes}
    \vspace{-0.2 in}
\end{figure}

Each map was tested under 4 input conditions. As a baseline, the Hierarchical CPOMDP policy was implemented without human input, with the cop relying only on it's visual sensor to gather information about the world. Second, the policy was implemented with a human who didn't respond to the robot pull questions, and only provided human push statements at their own discretion. Third, the policy received a human who only responded to robot pull questions, and ignored human push. Finally, the policy was implemented with a human who used both the robot pull questions and human push statements to give information. The resulting times required to catch the robber are summarized in Fig. \ref{fig:allCatchTimes}.

\subsection{The Familiar Map}

Across each input condition, tests were run with the robber's initial position in 3 different rooms, the Library, the Study, and the Kitchen. The cop's initial position was constant throughout all tests as the far right end of the hallway. The cop's also held an identical initial belief for each test, with belief dispersed equally between rooms. 

The tests showed that across starting positions, either the "Only Robot Pull", or "Both Robot Pull and Human Push" input conditions tended to require less time to catch the robber than either the "Only Human Push" or "No Human" input conditions. This is expected as the policy was trained assuming it would be able to pull information from the human, and the policies questions should logically be the most valuable to the cop for a given belief. Furthermore, the "Only Human Push" input condition universally improved on the times for the "No Human" input condition, demonstrating that  unexpected 
human data can be useful. 

\begin{figure}[t]
\centering	\includegraphics[width=0.35\textwidth]{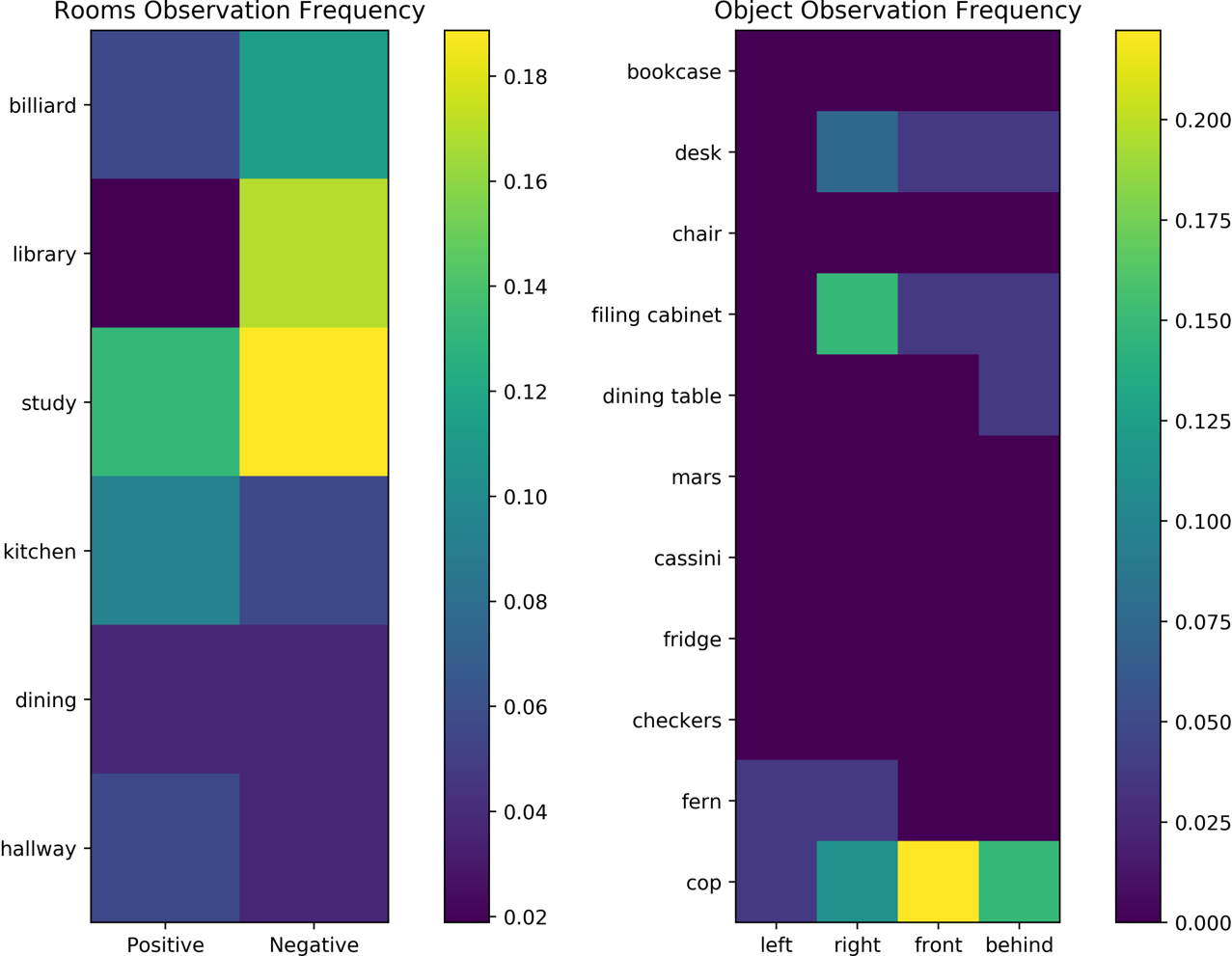}
	\caption{\scriptsize Heatmap of observations for the first map.}
    \label{fig:obsMapA}
    \vspace{-0.15 in}
\end{figure}

Over all tests in the first map 79 observations were given, averaging approximately 9 human inputs per test excluding the "No Human" condition. About 53\% of all statements were positive relations, eg. "I know the Robber is in the Study". Limited to observations about rooms, the human observer only gave positive observations 40\% of the time. When referencing objects 78\% of observations were positive. The human referenced rooms about twice as much as they did objects, as shown in Figure \ref{fig:obsMapA}. 

\begin{figure}[t]
\centering	\includegraphics[width=0.45\textwidth]{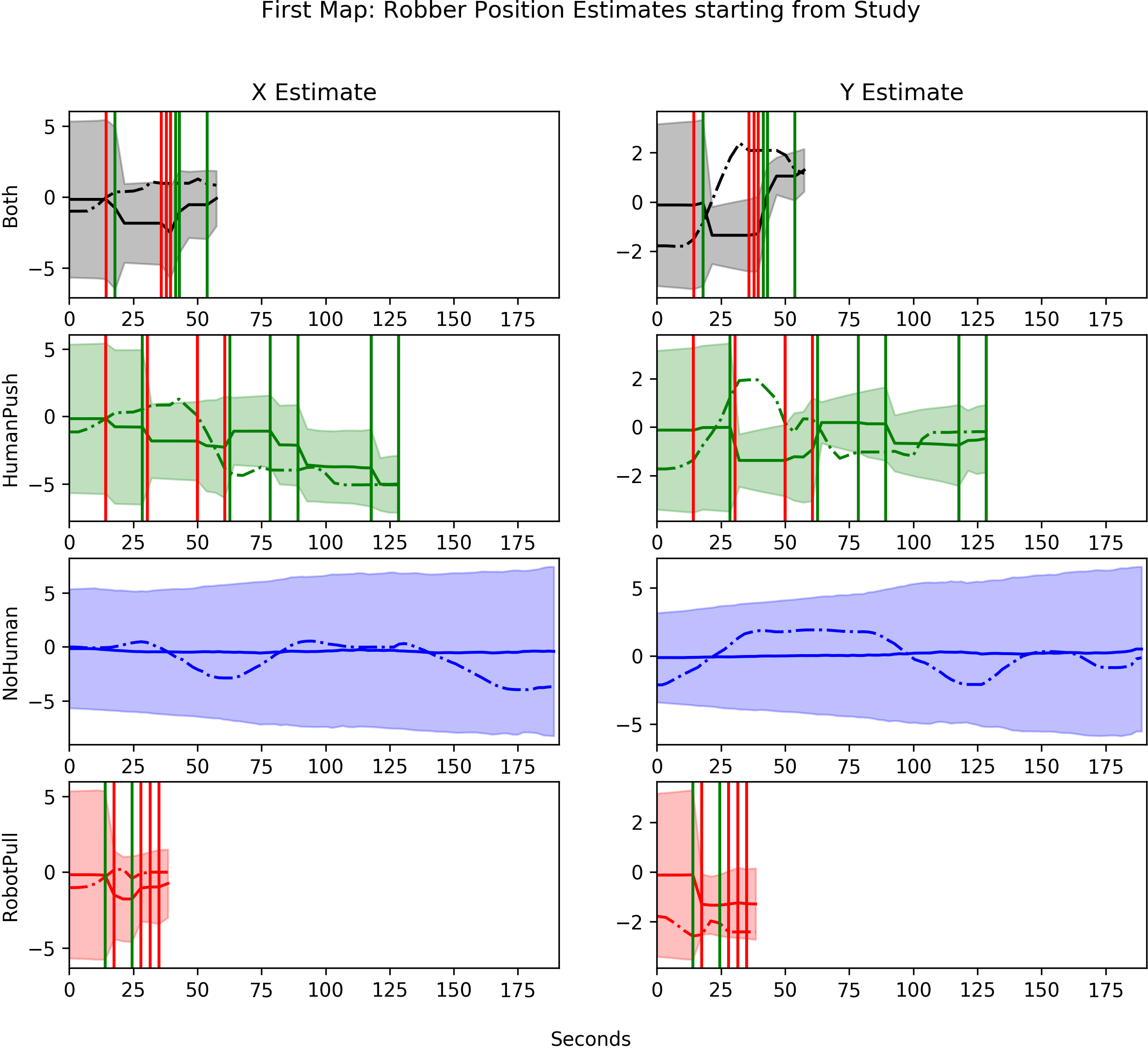}
	\caption{\scriptsize Summary of cop's beliefs for the first map. Mean and 2-sigma bounds of the cop's belief are plotted against robber's true position (dashed line). Vertical lines are color coded for positive (green) and negative (red) human statements.}
    \label{fig:robberEstMapA}
    \vspace{-0.1 in}
\end{figure}

The cop's beliefs are summarized in Figure \ref{fig:robberEstMapA} for the 4 test runs with the robber starting in the Study. For each input type, the mixture mean and 2-sigma bounds are plotted along with the robber's actual position. The "No Human" input condition sees the belief expanding faster than the cop's viewcone can clear it, as the robber dynamics are taken into account at each step. Human observations, shown as vertical lines in the plot, can cause dramatic shifts. The robber's position is can be seen to be generally well bounded by the cop's belief, which can correct for errors through additional human observations. 

\subsection{The Unfamiliar Map}

For the second map, the human observer was familiar with the task and platform, but not with the map itself, shown in Figure \ref{fig:layoutMapC}. The locations of the rooms and positions of the objects within were kept unfamiliar until the beginning of testing. As in the first map, 4 input conditions were tested over multiple initial robber positions, in this case the Billiard Room and the Study. Across all tests, the cop's initial position was set in the Kitchen, and the belief was evenly distributed between rooms. 
%
%
The timing results from the test, summarized in Figure \ref{fig:timesMapC}, are generally comparable with those of the first map, taking an additional 11 seconds to catch the robber on average. The comparison between input conditions also remains consistent, with the unfamiliar map results even suggesting an additional advantage for the "Both" condition over "Robot Pull Only". 

\begin{figure}[t]
\centering	\includegraphics[width=0.35\textwidth]{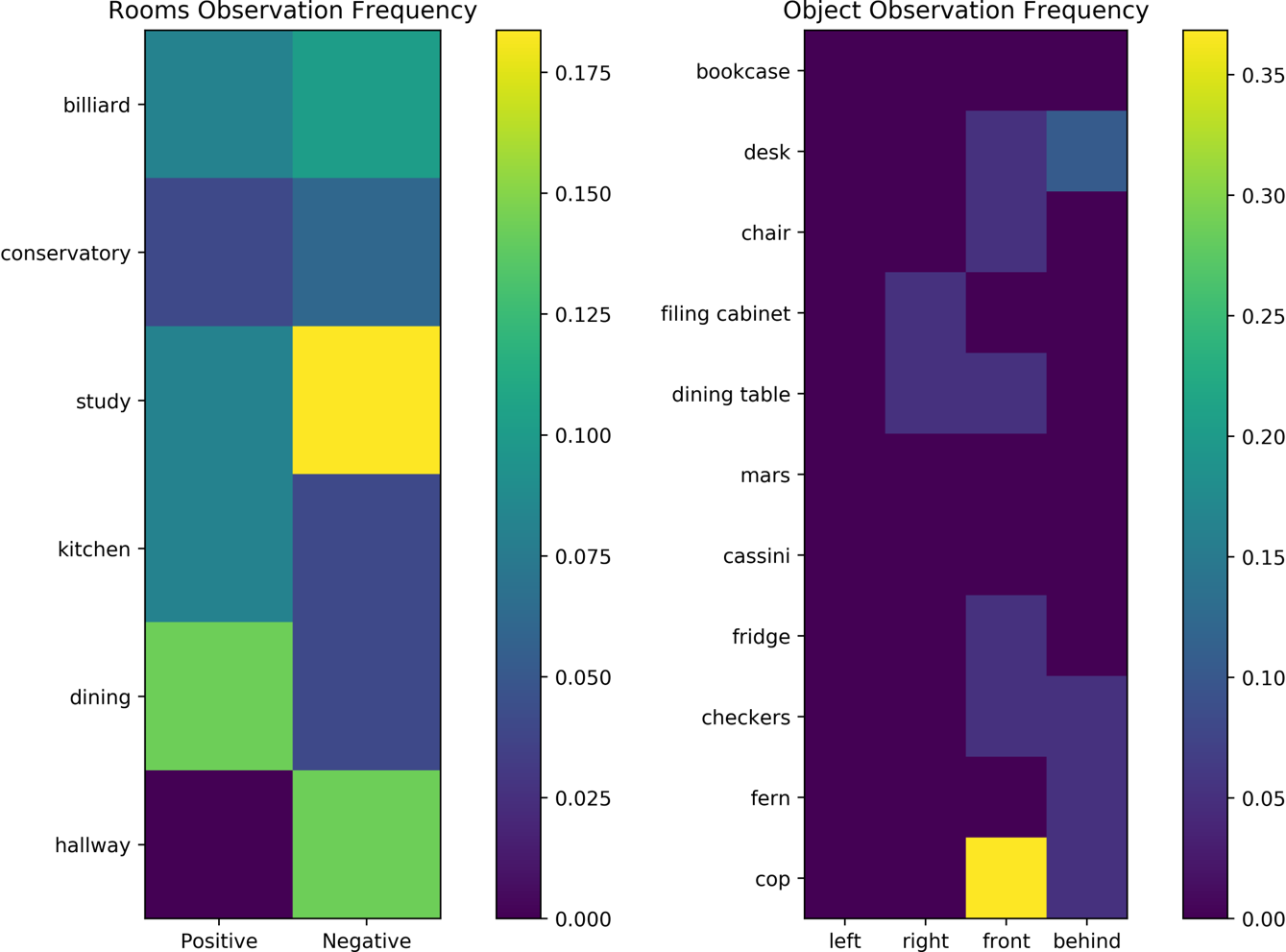}
	\caption{\scriptsize Heatmap of observations for the second map.}
    \label{fig:obsMapC}
    \vspace{-0.2 in}
\end{figure}
\begin{figure}[t]
\centering	\includegraphics[width=0.45\textwidth]{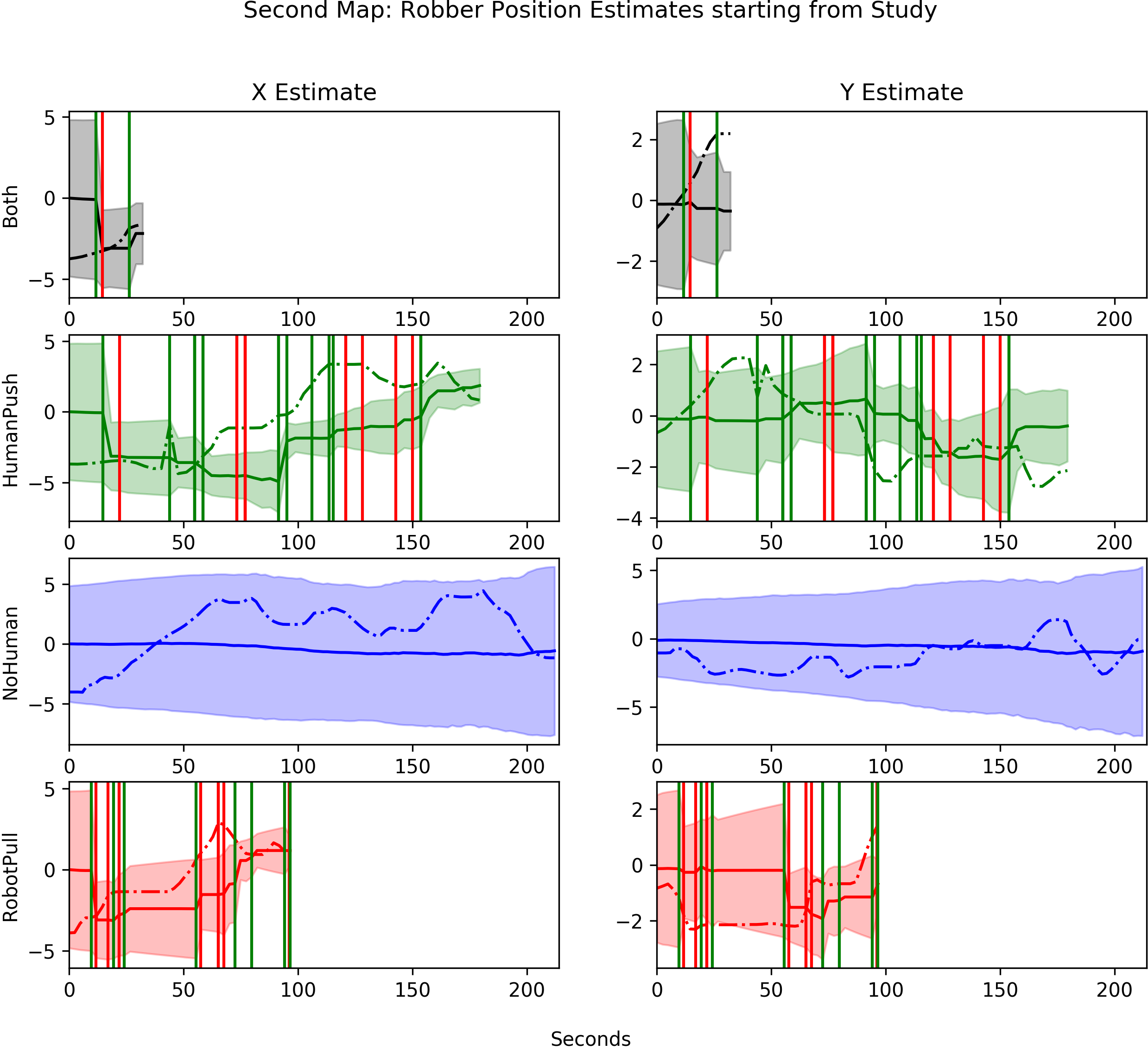}
	\caption{\scriptsize Summary of cop's beliefs for the second map. Mean and 2-sigma bounds of the cop's belief are plotted against robber's true position (dashed line). Vertical lines are color coded for positive (green) and negative (red) human statements. As expected, the unfamiliar environment leads to less accurate beliefs in the Human Push scenario.}
    \label{fig:robberEstMapC}
    \vspace{-0.25 in}
\end{figure}

For all tests in the second map there were a total of 66 observations, with an average of 11 human inputs per test excluding the "No Human" condition. In this case about 47\% of all statements were positive relations, with 42\% positives for room observations and 58\% positive for objects. Rooms were referenced almost 3 times as much as objects, with frequencies for each statement shown in Figure \ref{fig:obsMapC}. 
%
The cop's beliefs, summarized in Figure \ref{fig:robberEstMapC}, are once again a reasonable estimate of the robber's position despite slightly more errors. 

\subsection{Discussion}

\begin{figure*}[t!]
      \centering	
      \begin{subfigure}[t]{.45\textwidth}
          \includegraphics[width=\textwidth]{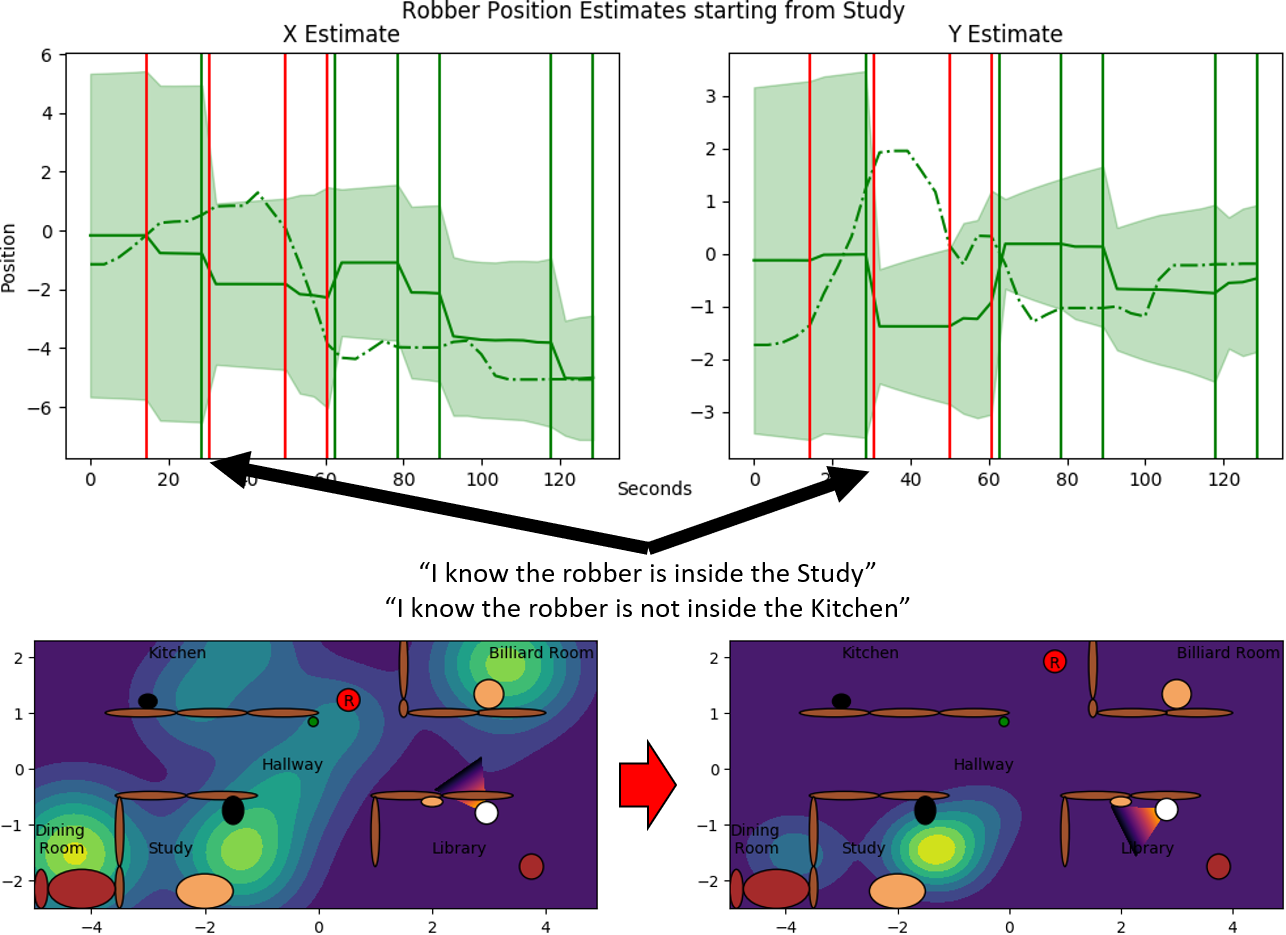}
          \caption{\scriptsize Human gives inaccurate series of observations}
          \label{fig:humanWrong}
      \end{subfigure}
      ~
      \begin{subfigure}[t]{.45\textwidth}
          \includegraphics[width=\textwidth]{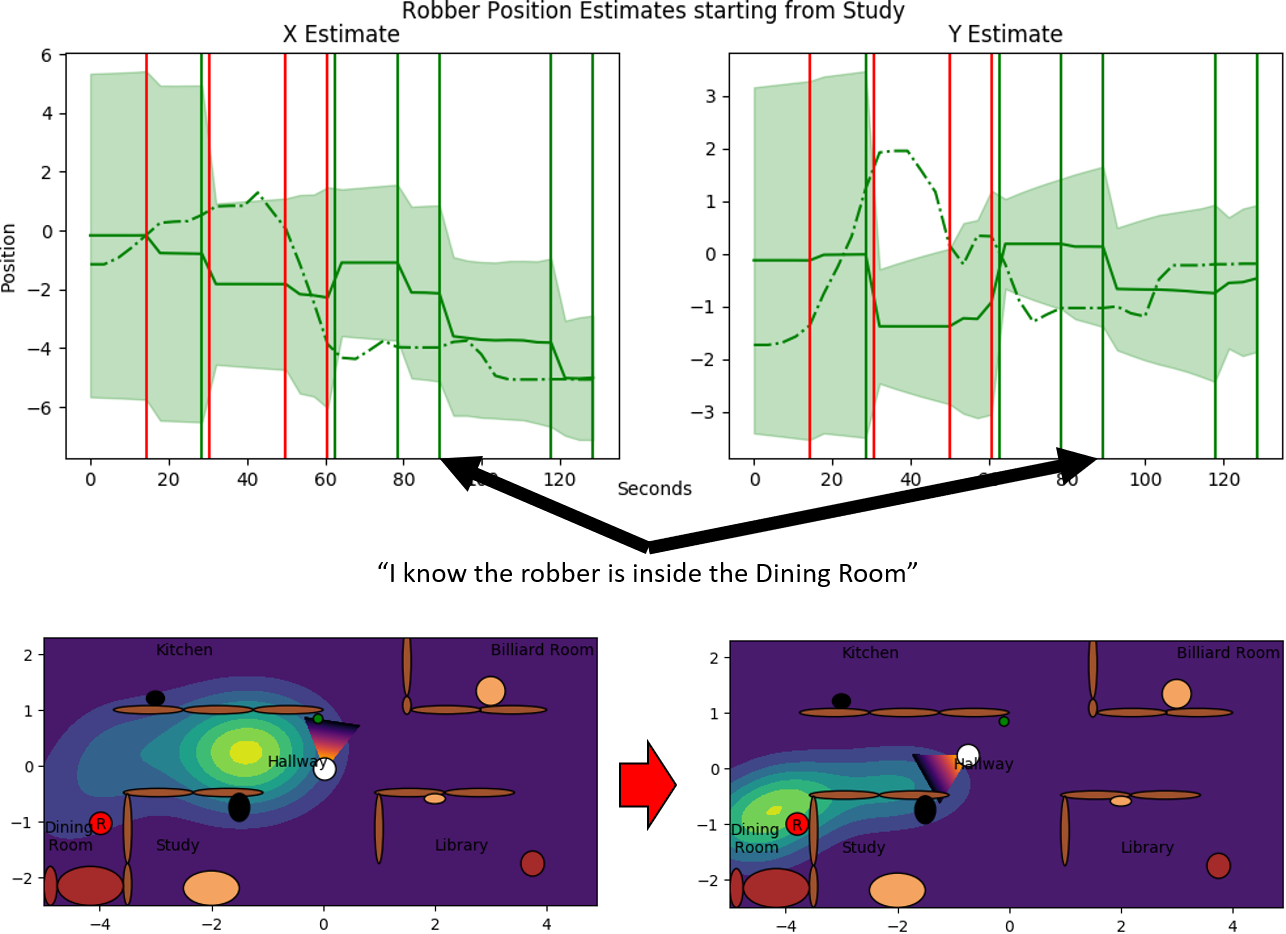}
          \caption{\scriptsize Human shifts belief with information unavailable to the cop}
          \label{fig:humanHelpful}
      \end{subfigure}
      \caption{\scriptsize Left: The human gives a series of mistaken observations. Right: The human gives a helpful statement}
      \label{fig:storyBoard}
      \vspace{-0.2 in}
\end{figure*}

The cop using the Hierarchical CPOMDP approach succeeded in all cases at catching the robber, and was demonstrably quicker in cases where it received and fused human information. Of particular note is the improvement of the "Robot Pull Only" input condition over the "Human Push Only" condition. This implies that information delivered at the policy's request was more valuable than that which the human decided to volunteer. As the policy is meant to approximate the optimal value function for the problem, this serves as evidence of its efficacy.

The system was also able to adapt to false information from the human sensor, as displayed in Figure \ref{fig:storyBoard}. In Figure \ref{fig:humanWrong}, after the robber passed in front of the security camera in the Study while moving into the Kitchen, the human unintentionally gave a series of false observations, rapidly shifting the belief from an uncertain but reasonable one to one that was decidedly inaccurate. Later in the same run, the human was able to combine visual information from both the Hallway camera and the cop's viewcone to indicate correctly that the robber had moved into the Dining Room, as shown in Figure \ref{fig:humanHelpful}.  

\begin{figure}
      \centering	
      \begin{subfigure}[t]{.35\textwidth}
          \includegraphics[width=\textwidth]{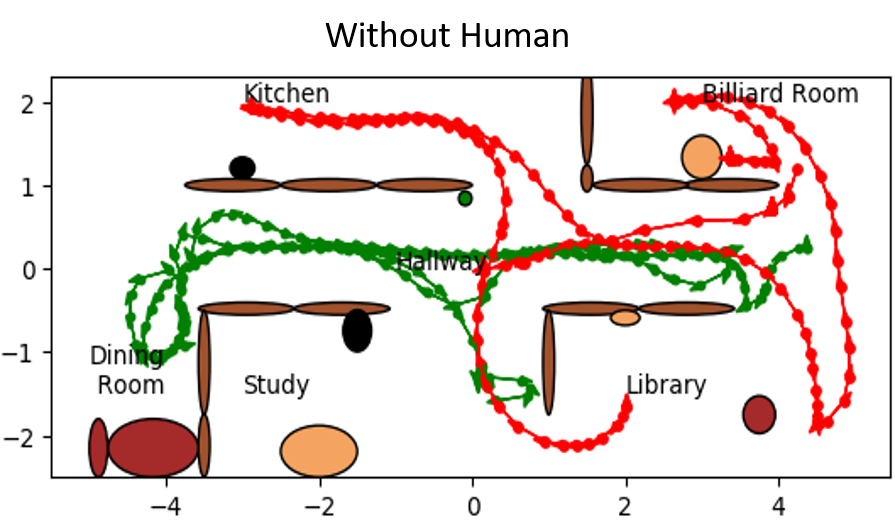}
          \caption{\scriptsize Example paths for "No Human" input condition}
          \label{fig:pathCompNone}
      \end{subfigure}
      
      \begin{subfigure}[t]{.35\textwidth}
          \includegraphics[width=\textwidth]{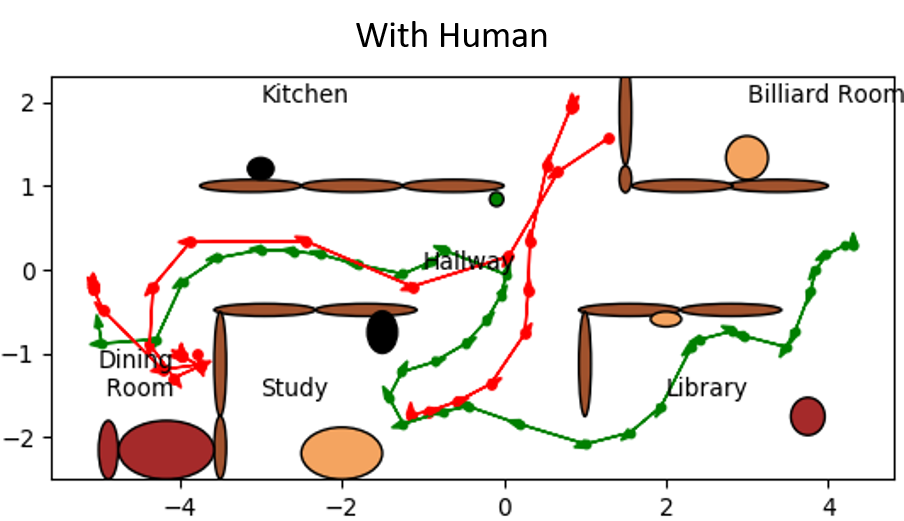}
          \caption{\scriptsize Example paths for "Human Push Only" input condition}
          \label{fig:pathCompHuman}
      \end{subfigure}
      \caption{\scriptsize Cop (green) and robber (red) paths without vs. with human sensor input.}
      \label{fig:pathComps}
      \vspace{-0.1 in}
\end{figure}

With human information, the policy was able to direct the cop more efficiently. As shown in Figure \ref{fig:pathCompNone}, without any human sensor data the policy primarily directs the cop to patrol the Hallway, popping in and out of individual rooms along the way. This behavior is reasonable considering the Hallway's position as a hub room, where the cop could expect to eventually stumble upon the robber as it moves from room to room. This displays the robustness of the policy's action selection in the absence of expected information. However, when a human operator is able to provide information as in Figure \ref{fig:pathCompHuman}, the policy chooses a path through the Library, and ends up tracking the robber directly through the Study, and into the Hallway, finally cornering it in the Dining Room.

\section{Conclusion}
%
We developed and validated a novel collaborative human-machine sensing solution for dynamic target search. Our approach used continuous partially observable Markov decision process (CPOMDP) planning to generate vehicle trajectories that optimally exploit imperfect detection data from onboard sensors and semantic natural language observations that can be requested from human sensors. The main innovation was a scalable hierarchical Gaussian mixture model formulation for efficiently solving CPOMDPs with semantic observations in continuous dynamic state spaces. The approach was demonstrated with a real human-robot team engaged in dynamic indoor target search and capture scenarios on a custom testbed. The results showed that combined human-robot sensing not only enhances target localization quality (as expected), but that the resulting CPOMDP policies provide sensible simultaneous search movements and semantic human sensor queries that allow the search vehicle to intercept the target more efficiently. The resulting CPOMDP policies are robust and effective even with irregular/unpredictable inputs and occasional errors from the human sensor. 

Ongoing and future research will focus on semantic data fusion in problems where we relax our assumptions of: known number of targets; known search environment/map and semantic reference objects; known search vehicle states; and known human sensor parameters. 
These problems are significantly more challenging to solve, but also have important practical implications for applications involving target search in highly uncertain environments, e.g. search and rescue or disaster relief. 
Building on the work here and in \cite{Burks2017}, we will investigate how semantic human sensor data can be actively leveraged for online interactive learning and planning, as well as online state estimation and perception.



\bibliographystyle{IEEEtran}
\bibliography{AhmedRefs,moreExtraRefs}

\begin{thebibliography}{10}
\providecommand{\url}[1]{#1}
\csname url@samestyle\endcsname
\providecommand{\newblock}{\relax}
\providecommand{\bibinfo}[2]{#2}
\providecommand{\BIBentrySTDinterwordspacing}{\spaceskip=0pt\relax}
\providecommand{\BIBentryALTinterwordstretchfactor}{4}
\providecommand{\BIBentryALTinterwordspacing}{\spaceskip=\fontdimen2\font plus
\BIBentryALTinterwordstretchfactor\fontdimen3\font minus
  \fontdimen4\font\relax}
\providecommand{\BIBforeignlanguage}[2]{{%
\expandafter\ifx\csname l@#1\endcsname\relax
\typeout{** WARNING: IEEEtran.bst: No hyphenation pattern has been}%
\typeout{** loaded for the language `#1'. Using the pattern for}%
\typeout{** the default language instead.}%
\else
\language=\csname l@#1\endcsname
\fi
#2}}
\providecommand{\BIBdecl}{\relax}
\BIBdecl

\bibitem{Kaupp2007}
T.~Kaupp, B.~Douillard, F.~Ramos, A.~Makarenko, and B.~Upcroft, ``{Shared
  Environment Representation for a Human-Robot Team Performing Information
  Fusion},'' \emph{Journal of Field Robotics}, vol.~24, no.~11, pp. 911--942,
  2007.

\bibitem{Bourgault2008}
F.~Bourgault, A.~Chokshi, J.~Wang, D.~Shah, J.~Schoenberg, R.~Iyer, F.~Cedano,
  and M.~Campbell, ``{Scalable Bayesian human-robot cooperation in mobile
  sensor networks},'' in \emph{International Conference on Intelligent Robots
  and Systems}, 2008, pp. 2342--2349.

\bibitem{Khaleghi10}
B.~Khaleghi, A.~Khamis, and F.~Karray, ``Random finite set theoretic based
  soft/hard data fusion with application for target tracking,'' in \emph{2010
  Conf. on Multisensor Fusion and Integration for Intelligent Sys. (MFI 2010)},
  Salt Lake City, 2010, pp. 50--55.

\bibitem{Dani2014}
A.~Dani, M.~McCourt, J.~Curtis, and S.~Mehta, ``Information fusion in
  human-robot collaboration using neural network representation,'' in
  \emph{2014 IEEE Int'l Conf. on Systems, Man and Cybernetics}.\hskip 1em plus
  0.5em minus 0.4em\relax IEEE, 2014, pp. 2114--2120.

\bibitem{Ahmed-TRO-2013}
N.~Ahmed, E.~Sample, and M.~Campbell, ``Bayesian multicategorical soft data
  fusion for human-robot collaboration,'' \emph{IEEE Trans. on Robotics},
  vol.~29, pp. 189--206, 2013.

\bibitem{Sweet2016}
N.~Sweet and N.~Ahmed, ``{Structured synthesis and compression of semantic
  human sensor models for Bayesian estimation},'' \emph{Proceedings of the
  American Control Conference}, vol. 2016-July, no.~2, pp. 5479--5485, 2016.

\bibitem{Burks2017}
L.~Burks and N.~Ahmed, ``Optimal continuous state pomdp planning with semantic
  observations,'' in \emph{2017 IEEE Conference on Decision and Control}.\hskip
  1em plus 0.5em minus 0.4em\relax IEEE, 2017, pp. 1509--1516.

\bibitem{HallBook}
D.~L. Hall and J.~M. Jordan, \emph{{Human-centered Information Fusion}}.\hskip
  1em plus 0.5em minus 0.4em\relax Artech House, 2010.

\bibitem{Runnalls-AES-2007}
A.~R. Runnalls, ``{Kullback-Leibler Approach to Gaussian Mixture Reduction},''
  \emph{IEEE Transactions on Aerospace and Electronic Systems}, vol.~43, no.~3,
  pp. 989--999, 2007.

\bibitem{Koch04}
W.~Koch, ``On "negative" information in tracking and sensor data fusion:
  Discussion of selected examples,'' in \emph{FUSION 2004}, 2004.

\bibitem{Kaupp2010}
\BIBentryALTinterwordspacing
T.~Kaupp, A.~Makarenko, and H.~Durrant-Whyte, ``{Human�robot communication
  for collaborative decision making � A probabilistic approach},''
  \emph{Robotics and Autonomous Systems}, vol.~58, no.~5, pp. 444--456, May
  2010. [Online]. Available:
  \url{http://linkinghub.elsevier.com/retrieve/pii/S0921889010000400}
\BIBentrySTDinterwordspacing

\bibitem{Pineau2003}
J.~Pineau, G.~Gordon, and S.~Thrun, ``{Point-based value iteration: An anytime
  algorithm for POMDPs},'' \emph{IJCAI International Joint Conference on
  Artificial Intelligence}, pp. 1025--1030, 2003.

\bibitem{Spaan2005}
M.~T.~J. Spaan and N.~Vlassis, ``{Perseus: Randomized point-based value
  iteration for POMDPs},'' \emph{Journal of Artificial Intelligence Research},
  vol.~24, pp. 195--220, 2005.

\bibitem{Kurniawati2008}
H.~Kurniawati, D.~Hsu, and W.~S. Lee, ``{SARSOP: Efficient point-based POMDP
  planning by approximating optimally reachable belief spaces},'' \emph{Proc.
  Robotics: Science and Systems}, 2008.

\bibitem{Porta2006}
M.~S. P.~P. JM~Porta, N~Vlassis, ``{Point-based value iteration for continuous
  POMDPs},'' \emph{IJCAI International Joint Conference on Artificial
  Intelligence}, vol.~7, pp. 1968--1974, 2006.

\bibitem{Brunskill2010}
E.~Brunskill, L.~P. Kaelbling, T.~Lozano-Perez, and N.~Roy, ``{Planning in
  partially-observable switching-mode continuous domains},'' \emph{Annals of
  Mathematics and Artificial Intelligence}, vol.~58, no.~3, pp. 185--216, 2010.

\end{thebibliography}

\end{document}